\documentclass[accepted]{uai2025} 

\newcommand{\Mc}{\mathcal{M}}
\newcommand{\Xc}{\mathcal{X}}

\newcommand{\R}{\mathbb{R}}

\usepackage{booktabs} 
\usepackage{makecell} 

\usepackage[american]{babel}
\usepackage{makecell}

\usepackage{natbib} 
\usepackage{times}

\title{Automated Manifold Learning for Reduced Order Modeling}

\author{ {\bf Imran Nasim$^{1,2}$ and Melanie Weber$^3$} \\
$^1$IBM UK, $^2$Department of Mathematics, University of Surrey, $^3$School of Engineering and Applied Sciences, Harvard University\\
Contact: imran.nasim@ibm.com, mweber@seas.harvard.edu}

\usepackage{caption} 
\usepackage{color} 
\usepackage{graphicx} 

\usepackage{algorithm}
\usepackage{algorithmic}

\usepackage{microtype} 
\usepackage{booktabs}  

\usepackage{amsmath}
\usepackage{amsthm}
\usepackage{amssymb}
\usepackage{mathtools}

\begin{document}

\maketitle

\begin{abstract}
The problem of identifying geometric structure in data is a cornerstone of (unsupervised) learning. As a result, Geometric Representation Learning has been widely applied across scientific and engineering domains. 
In this work, we investigate the use of Geometric Representation Learning for the data-driven discovery of system dynamics from spatial-temporal data. We propose to encode similarity structure in such data in a \emph{spatial-temporal proximity graph}, to which we apply a range of classical and deep learning-based manifold learning approaches to learn reduced order dynamics. We observe that while manifold learning is generally capable of recovering reduced order dynamics, the quality of the learned representations varies substantially across different algorithms and hyperparameter choices. This is indicative of high sensitivity to the inherent geometric assumptions of the respective approaches and suggests a need for careful hyperparameter tuning, which can be expensive in practise. To overcome these challenges, we propose a framework for \emph{Automated Manifold Learning}, which selects a manifold learning approach and corresponding hyperparameter choices based on representative subsamples of the input graph. We demonstrate that the proposed framework leads to performance gains both in scalability and in the learned representations' accuracy in capturing local and global geometric features of the underlying system dynamics.
\end{abstract}

\section{Introduction}
    The problem of learning the dynamics of a complex system from spatial-temporal data arises in applications across the Sciences and Engineering. Often such data has a low-dimensional structure, usually induced by inherent symmetries in the underlying system, which arise from fundamental laws of physics. In the dynamical systems literature there is a growing body of work on learning dynamics from data, e.g.,~\citep{lusch2018deep,otto_linearly_2019,brunton2016discovering,fasina_geometric_nodate,nasim_JLS_scml}, albeit without explicitly encoding geometric assumptions. A related body of literature considers the problem of identifying and characterizing low-dimensional structure in data, ranging from classical manifold learning approaches~\citep{kruskal1964multidimensional,tenenbaum2000global,belkin2003laplacian} to more recent deep learning based methods~\citep{Deepwalk_perozzi_2014,grover2016node2vec,Kipf_gcn_2017,nasim2024_dynamicalAE,scml}. Manifold learning approaches are unsupervised methods, which can be applied to small, imperfect data and whose simplicity allows for learning transparent and interpretable data representations. These methods are typically applied to static (non-time dependent) data and require careful hyperparameter tuning. The question of which method performs best in a given setting is highly data- and domain-dependent. 

In this work, we investigate the utility and model-dependency of such manifold learning approaches for the recovery of reduced order dynamics from \emph{spatial-temporal} data. We consider data that is generated by physical systems whose ground truth dynamics are characterized by solutions to Partial Differential Equations (PDEs). We propose a framework for recovering the system dynamics in a purely unsupervised, data-driven manner, relying on a spatial-temporal proximity graph that encodes similarity structure in spatial-temporal observations of the system. We perform a systematic qualitative and quantitative analysis of the different approaches with the aim of identifying methods that can reliably recover the reduced order dynamics. Our analysis uncovers significant differences between the methods in terms of their ability to capture local and global geometric features in spatial-temporal data. In particular, we observe that the quality of the learned representations is highly sensitive to the inherent geometric assumptions of the different approaches and the choice of hyperparameters. This sensitivity could amplify challenges arising in the presence of noise or partial observations, as well as when combining data from heterogeneous sources; all of which we encounter frequently when modeling complex dynamical systems in practice.
While there have been significant recent advances in addressing these challenges in other reduced order methods~\citep{fung2016_dynamics, champion2019_partialdata, bakarji2023_discovering}, the proposed solutions do not translate easily to the manifold learning setting. 

Fine-tuning design choices in manifold learning approaches can be prohibitively time-consuming and prevent scalability, often rendering manifold learning approaches inapplicable in practice. To overcome these shortcomings, we propose an \emph{automated manifold learning} framework, which allows for selecting the most suitable manifold learning approach and corresponding hyperparameters through optimization over small, representative subsamples of the spatial-temporal proximity graph. We demonstrate through numerical experiments on synthetic and real data  that the proposed approach leads to performance gains both in terms of scalability and in the learned representation’s accuracy in recovering the underlying system dynamics.

\paragraph{Related work.} \emph{Manifold Learning} is a family of classical algorithms for Geometric Representation Learning, notable instances include \textsc{Isomap}~\citep{tenenbaum2000global}, \textsc{MDS}~\citep{kruskal1964multidimensional}, \textsc{LLE}~\citep{roweis2000nonlinear} and Laplacian Eigenmaps~\citep{belkin2003laplacian}. Recently, graph-based approaches, such as \textsc{DeepWalk}~\citep{Deepwalk_perozzi_2014}, \textsc{node2vec}~\citep{grover2016node2vec}, as well as Graph Neural Networks~\citep{Kipf_gcn_2017,hamilton} have been applied in the same setting. However, the performance of these approaches varies widely in practice; to the best of our knowledge, a systematic comparison of trade-offs between those methods is largely missing in the literature. A growing body of literature studies the problem of \emph{learning dynamics from data}, including via Deep Learning~\citep{lusch2018deep}, Autoencoders~\citep{otto_linearly_2019}, Dynamic Mode Decomposition~\citep{brunton2016discovering} and Neural PDEs~\citep{fasina_geometric_nodate}, albeit without explicitly encoding geometric assumptions. A Neural PDE-based approach proposed by~\citet{fasina_geometric_nodate} is the closest to the approach proposed in this work, as it utilizes a similar spatial-temporal similarity graphs as constructed in our framework.
\emph{Automated Machine Learning} (short \emph{AutoML}) seeks to determine the best model, including architecture choices, training protocols and hyperparameters. AutoML has been applied in various settings, including classification and regression on tabular data~\citep{yang2022tabnas}, matrix and tensor factorization~\citep{yang2020efficient}, as well as Graph Machine Learning~\citep{tu2019AutoNE,zhang2021_automlgraph_survey}, among others. However, to the best of our knowledge, AutoML has not yet been applied to manifold learning.

\paragraph{Summary of Contributions.}
The main contributions in this work are three-fold:
\begin{enumerate}
    \item We propose a simple, unsupervised framework for learning reduced order dynamics from spatial-temporal data via manifold learning. A key component of our approach is a spatial-temporal similarity graph to which we apply classical and deep learning-based manifold learning approaches. 
    \item A systematic study of different manifold learning algorithms indicates that the quality of the learned representations is highly sensitive to model and hyperparameter choices.
    \item We propose an \emph{Automated Manifold Learning} framework for optimizing model selection and hyperparameter tuning, yielding significant performance gains and improved scalability.
\end{enumerate}

\section{Background}
\subsection{Learning manifolds from data}
In this work, we assume that data lies on or near a low-dimensional manifold, an assumption commonly known as the \emph{manifold hypothesis}.  A large body of literature has evolved around the problem of reconstructing manifolds from data (\emph{manifold learning}). Formally, let $\Xc \times (0,T) \subseteq \mathbb{R}^{d+1}$ denote time series data representing observations of a physical system. In practise, these measurements may be inaccurate (e.g., affected by noise) or incomplete (e.g., not representative of the low-dimensional manifold). We want to learn a low-dimensional representation of the system's dynamics, which allows for characterizing its local and global geometric properties, characterized by a \emph{representation function} $\phi: \Xc \times (0,T) \rightarrow \Mc$, where $\Mc \subset \R^{d+1}$ (with dimension $\dim(\Mc) \ll d$). We will leverage both classical~\citep{kruskal1964multidimensional,tenenbaum2000global,roweis2000nonlinear} and deep learning based manifold learning~\citep{Deepwalk_perozzi_2014,Kipf_gcn_2017,Velickovic_GATS_2017} for  approximations of the image of the representation function $\phi(\Xc)$. We will describe and implement all manifold learning approaches in terms of common building blocks, which we will introduce below. 

\paragraph{Spatial-temporal Similarity Graph.}
A key component of every representation learning algorithm is a \emph{similarity measure} imposed on the data space, which provides a means for quantifying the similarity or dissimilarity of pairs of data points. In the algorithms considered here, we encounter three types of similarity measures: First, \emph{Euclidean distances} in the high-dimensional observation space, which we can directly measure. Second, \emph{geodesic distances}, which encode "true" distances on the low-dimensional data manifold. We cannot measure these distances directly, but assume that they can be approximated from pairwise shortest-path distances on the similarity graph. And third, \emph{probabilistic distances}, which are local approximations of geodesic distances, determined via the likelihood that two nodes co-occur in a random walk on a similarity graph. The choice of similarity measure is often not unique and represents a crucial design choice. A key subroutine of many similarity measures is the construction of a \emph{similarity graph}. Unlike in the typical manifold learning setting, our graph is \emph{time-dependent}, with nodes representing time points; the spatial discretization is encoded in the node attributes. To construct a \emph{spatial-temporal similarity graph}, nodes are connected by edges, if they are $k$-nearest neighbors with respect to the similarity of the node attributes. The construction requires a careful choice of the parameters $k$, which often requires a grid search in practise. 

\paragraph{Embedding.} Assuming access to a similarity measure, we can compute a point configuration on a low-dimensional manifold. This process implicitly approximates a representation function $\phi: \Xc \times (0,T) \rightarrow \Mc$, which embeds the data $\Xc$ into the target manifold $\Mc$. In classical manifold learning methods, the point configuration is usually computed by solving an eigenvalue problem. Shallow graph embeddings are optimization-based, i.e., a loss function, which is defined based on the chosen similarity measure, is optimized to compute a point configuration. In deep learning-based embeddings, point configurations are trained via message-passing.

\paragraph{Evaluation.} Once a point configuration is computed, we evaluate its utility. We propose to measure the quality of learned representations qualitatively via visualization and quantitatively via local and global Q-scores, which are classical measures of embedding quality~\citep{lee_scale-independent_2010}. We defer all formal definitions to Apx.~\ref{sec:q-scores}.

\subsection{Manifold Learning Algorithms}
In this section we briefly review representation learning algorithms considered in this study. A taxonomy is given in Table.~\ref{tab:mf-taxonomy} in Apx.~\ref{sec:app-methods}.

\paragraph{Classical Manifold Learning algorithms.}
Arguably the simplest approach for inferring low-dimensional structure in data is via linear projections, e.g., using \textsc{PCA}. The most common approach for learning non-linear subspaces from data are manifold learning algorithms. At the heart of those algorithms lies often an eigenvalue problem, where utilizing the eigenvectors associated with the top eigenvalues leads to projections that preserve pairwise distances globally (\emph{global methods}), whereas projections based on the eigenvectors corresponding to the bottom eigenvalues preserves local pairwise distances (\emph{local methods}). 
Examples of global methods are \textsc{Isomap}~\citep{tenenbaum2000global} and \textsc{MDS}~\citep{kruskal1964multidimensional}. We further consider \textsc{LLE}~\citep{roweis2000nonlinear} and \textsc{SE}~\cite{shi2000_SE} as examples of local methods. 

\paragraph{Deep Learning on Point Clouds.} 
One of the first Representation Learning approaches that utilized probabilistic similarity measures was \textsc{DeepWalk}\citep{Deepwalk_perozzi_2014}, a shallow node embedding method. 
More recently, approaches based on Graph Neural Network (GNN) architectures have been utilized for graph embeddings. Here, we consider \textsc{GraphSage}~\citep{hamilton}, \textsc{GCN}~\citep{Kipf_gcn_2017}  and \textsc{GAT}~\citep{Velickovic_GATS_2017}. 
 \textsc{GCN} and \textsc{GraphSage} are early GNN architectures, which extend convolutional neural networks to graph domains. \textsc{GAT}~\citep{Velickovic_GATS_2017} implements the idea of graph attention,  i.e., it translates the idea of \emph{transformers} to the graph setting. More details on all four approaches can be found in Apx.~\ref{sec:app-methods}.

 \subsection{Reduced-Order Modeling}
 Reduced-order modeling is a classical task in the study of complex dynamical systems, including systems whose dynamics can be modeled by partial differential equations. It aims to reduce the complexity of the system by identifying low-dimensional structure, which arises from dominant modes in the data. The reduced model can provide key insights into the underlying system and enable large-scale simulations and predictions that would be prohibitively expensive at full model size.
In comparison with recent deep learning based approaches for reduced-order modeling~\citep{lusch2018deep,fasina_geometric_nodate}, manifold learning is fully unsupervised, does not make prior assumptions on the geometry of the low-dimensional representations and can be more scalable.

\section{Methods}
\subsection{Automating Representation Learning}
\begin{figure*}[t]
    \centering
\includegraphics[width=\textwidth]{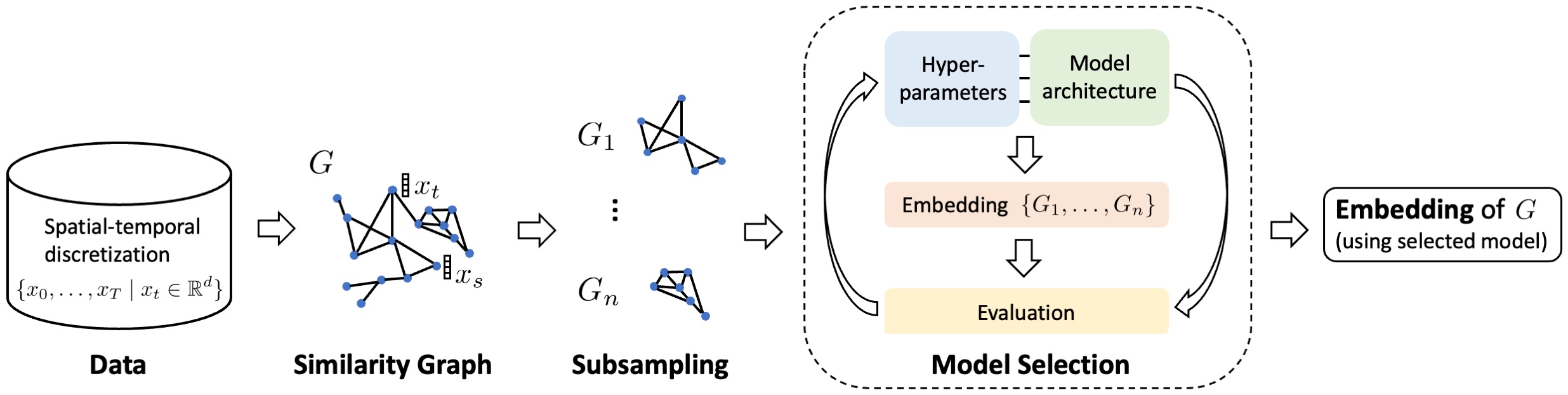}
    \caption{Overview of our Automated Representation Learning framework.}
    \label{fig:framework}
\end{figure*}
We begin by describing our automation pipeline (Fig.~\ref{fig:framework}). Its key components are a preprocessing step, which generates a set of subgraphs sampled from the input graph, as well as a model selection procedure, which determines the model architecture, training protocol and corresponding hyperparameters. In the following we assume that our input is a spatial-temporal proximity graph, constructed from given spatial-temporal observations of the underlying physical system.

\paragraph{Subsampling procedure.} Performing model selection directly on large input graphs is not computationally infeasible. Hence, we subsample the input graph by constructing a set of subgraphs that is representative of its topology. Our subsampling procedure is based on uniform random walks anchored at a randomly selected starting node. The subgraph consists of the set of nodes the random walk traverse; its connectivity is inherited from the original graph by restricting to the selected subset of nodes. 

\paragraph{Model selection procedure.}
Our model selection procedure consists of three key components, (i) meta-learning of geometric priors, (ii) automated hyperparameter tuning, and (iii) selection of the model architecture. Geometric priors include the dimension of the learned embedding (i.e., the feature dimension of the output) and its "curvature", in the sense of deciding whether the representation space is assumed to be linear or nonlinear. The latter is implicit in the choice of the embedding method. Important hyperparameters include the density of the spatial-temporal proximity graph ($k$) and the length of the random walk in the subsampling procedure (i.e., the size of the sampled subgraphs). Model selection encompasses the choice between linear and nonlinear dimensionality reduction, as well as the choice of classical or deep manifold learning methods. As part of the model selection we further choose a training protocol, which includes the choice of the optimizer and its learning rate. More details on the components of the model selection procedure and hyperparameter optimization can be found in Apx.~\ref{apx:model-select}.

\subsection{Evaluating Representation Quality}\label{sec:q-scores}
\emph{Q-scores}, originally introduced by~\citep{lee_scale-independent_2010}, are ranking losses which evaluate how well pairwise distances are preserved locally and globally by an embedding $\phi$. Q-scores are computed with respect to a \emph{co-ranking matrix} $Q=\{ q_{kl} \}_{k,l \in [N]}$, defined by
\begin{small}
\begin{align*}
    q_{kl} &:= \vert \{ (k,l): \; \mu_{ij}=k, \nu_{ij}=l\} \vert \\
    \mu_{ij} &:= \big| \big\{ k : \;
\big(d_\Xc(x_i, x_k) < d_\Xc(x_i, x_j) \; {\rm and} \\
&\hspace{2em} d_\Xc(x_i, x_k) = d_\Xc(x_i, x_j) \; {\rm if} \; 1 \leq k < j \leq N \big)
\big\} \big| \\
\nu_{ij} &:= \big| \big\{ k : \;
d_\Mc(x_i, x_k) < d_\Mc(\phi(x_i), \phi(x_j)), \\
&\hspace{2em} d_\Mc(\phi(x_i), \phi(x_k)) = d_\Mc(\phi(x_i), \phi(x_j)) \; \big\}_{1 \leq k < j \leq N} 
\big\} \big|
\end{align*}
\end{small}
As usual, $\vert I \vert$, denotes the cardinality of a set $I$. We define Q-curves as
\begin{equation*}
    Q(\phi,K) := \frac{1}{KN} \sum_{(k,l) \in I_K} q_{kl} \; ,
\end{equation*}
which are computed over blocks
$I_K := \{1, \dots, K \} \times \{1, \dots, K \}$
of the co-ranking matrix. Note that we only need to consider the upper half of $I_K$, due to symmetry. With respect to Q-curves, we compute local and global Q-scores of $\phi$:
\begin{align}
    K_{\max} &= \mathop{\arg\max}_{K} \left( Q(\phi,K) - \frac{K}{N-1} \right) \\
    Q_{\text{local}} (\phi) &= \frac{1}{K_{\max}} \sum_{K=1}^{K_{\max}} Q(\phi,K) \\
    Q_{\text{global}} (\phi) &= \frac{1}{N-K_{\max}} \sum_{K=K_{\max}}^{N-1} Q(\phi,K) \;.
\end{align}
Here, the local and global regimes are distinguished by a split of the Q-curve at $K_{\max}$. $Q_{local}$ and $Q_{global}$ attain values from 0 (worst) to 1 (best).

\subsection{Optimization Procedure}
To select the best manifold learning approach and the corresponding best hyperparameters, we evaluate the embedding quality of each method over the set of subgraphs with respect to Q-scores. We compare two different optimization procedures that utilize the taxonomy described above with manually fine-tuned baselines.

\paragraph{Random Search.}
Random search is a widely used technique for hyperparameter tuning. The key advantage of random search as compared to grid search is that the configuration search space is explored more efficiently, allowing for the consideration of a larger range of possible configurations~\citep{bergstra12a}. The first instance of our automated model selecting procedure uses a variant of random search.

\paragraph{Bayesian Optimization.}
Sequential model-based optimization (SMBO) \citep{hutter2011smbo} has been developed to leverage previous experiences and decrease the number of necessary trials. SMBO employs a surrogate function, derived from historical data, to estimate the unknown function that correlates a set of hyperparameters with their expected performance. One of the leading SMBO methods is Bayesian optimization \citep{snoek2012_bayesian}, which utilizes a Gaussian process to model the surrogate function. We also employ this approach which selects hyperparameters for subsequent trials by aiming to maximize expected improvement.

\section{Experiments}
\subsection{Experimental setup}
\begin{figure*}[ht]
    \centering
\includegraphics[width=0.95\textwidth]{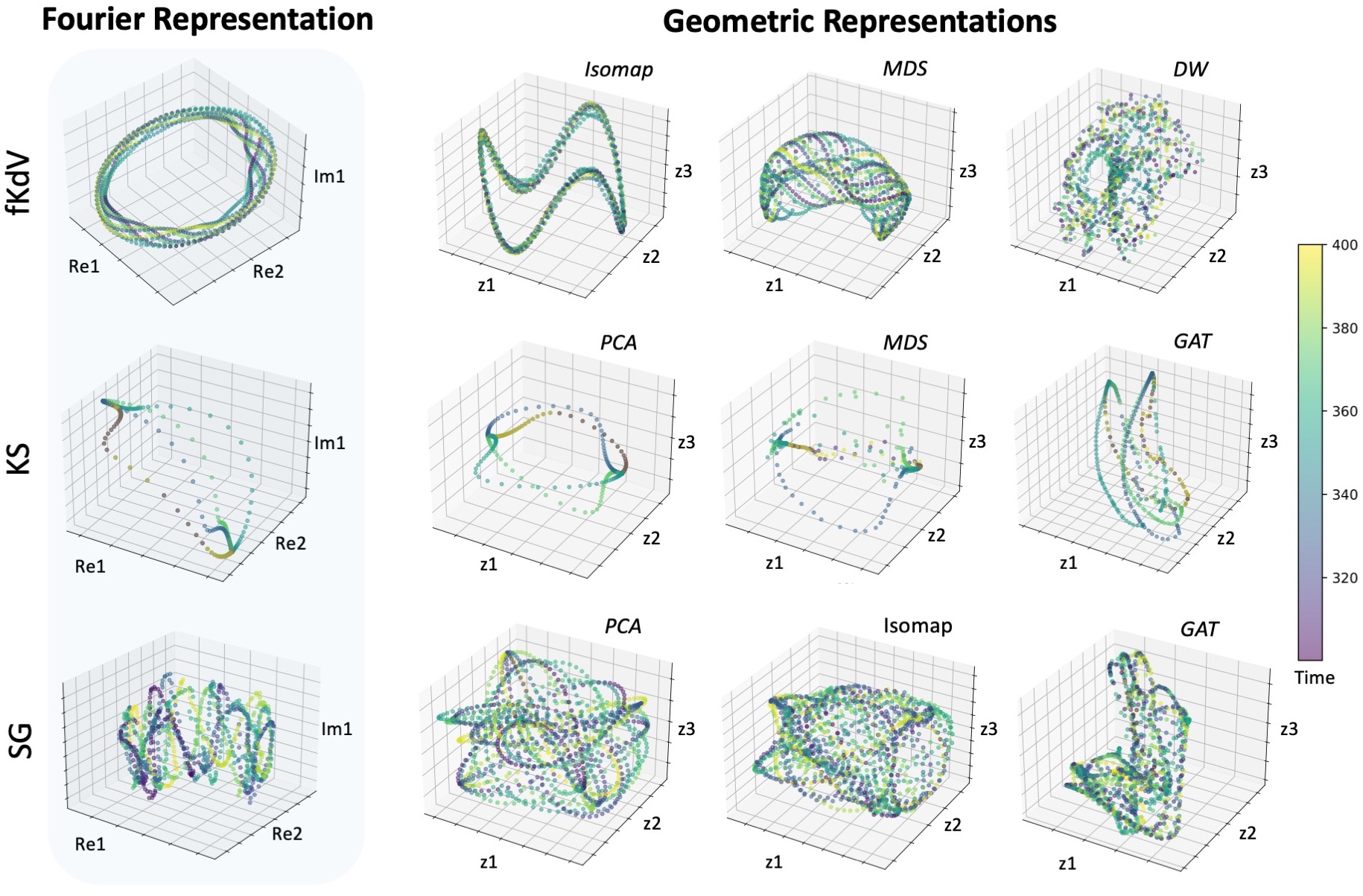}
    \caption{Three-dimensional visualizations of the reduced order dynamics (top row: forced Korteweg–de Vries (fKdV), middle row: Kuramoto–Sivashinksy (KS), bottom row: Sine Gordon (SG)). Left: Ground truth (virtual best) dynamics in Fourier space. Right: Geometric representations learned with top two classical and top deep-learning based approach. We see that classical methods outperform deep learning based methods. Among the classical methods, the best-performing ones are global embedding methods (PCA, MDS, Isomap) for all three systems. }
    \label{fig:results}
\end{figure*}

\paragraph{Case studies (Synthetic dynamic data).} The forced Korteweg–de Vries (fKdV) equation is an integrable non-linear PDE which is commonly used to model the weakly non-linear flow problem \cite{binder2005_fkdv,fkdv_2019}. In this study we consider the fKdV equation under the assumption of no disturbance where the equation can be written as~\cite{fkdv_2019}:
\begin{equation}\label{eq: fkdv}
    6u_t + u_{xxx} + (9u - 6(F-1))u_x = 0 \; .
\end{equation}
Here, $F$ is the depth-based Froude number. This model has been shown to exhibit both periodic travelling wave and soliton dynamics \cite{fkdv_2019}.

The second model we consider is the Kuramoto--Sivashinksy equation (KS).
The KS equation is a fourth-order non-integrable nonlinear PDE that can be used to model the pattern formation for several physical systems \cite{kuramoto_1978}. The viscous form of the equation can be written as 
\begin{equation}\label{KS-equation}
    u_{t} + uu_{x} + u_{xx} + \nu u_{xxxx} = 0,
\end{equation}
where $\nu$ is a coefficient of viscosity.
The KS equation is a highly complex model which captures the dynamics of spatio-temporal instabilities. The KS model also displays chaotic motion, and due to its non-integrability, gives rise to a rich array of solution types depending on the value of the viscosity parameter $\nu$. Most pertinent to this study, in the case where $\nu = \frac{16}{71}$, the KS equation exhibits bursting dynamics \citep{kirby1992_ks_bursting}.

In addition to the two models described above, we consider the Sine Gordon (SG) equation which is an integrable PDE. The SG equation gained popularity for the ability to exhibit soliton solutions \citep{hirota1972_SG} and can be written as
\begin{equation}\label{SG-equation}
    u_{tt} - u_{xx} + \sin{x} = 0.
\end{equation}

\paragraph{Data generation.} 
To obtain the dynamics data we simulated the fKdV and KS equation on a grid of 64 points on a periodic domain of $-\pi \leq x \leq \pi$. The Froude number in fKdV was assumed to be $F=1.5$. The initial wave profiles are of the form $u(x,0) = {A}\cos\left({k}x + \phi\right)$ where we fix $A=0.5$, $k=1$ and $\phi=1$. The trajectory data is collected from time $t=300$ with $\Delta t=0.1$ yielding 1000 time points, which captures the dynamics well after the initial transients have died out.
Numerical integrations of the fKdV and KS equations were performed using an explicit RK finite-difference scheme with a tolerance of $10^{-6}$ which we compared to a psuedo-spectral method to ensure accuracy. For the SG equation, we used a grid of 64 points and solved the model using an adaptive 5th order exponential time differencing method \citep{whalen2015_etd}.

\paragraph{Synthetic static data.} 
As a sanity check, we further test our framework on two synthetic data manifolds, the swiss roll and a sphere dataset, which forms a static, nonlinear manifold. We generate 1000 3-dimensional data points using the \textsc{sklearn} package with the default model configurations for the swiss roll and generate 1000 unique datapoints on the surface of a unit sphere. 

\begin{figure*}[t]
    \centering
\includegraphics[width=1.0\textwidth]{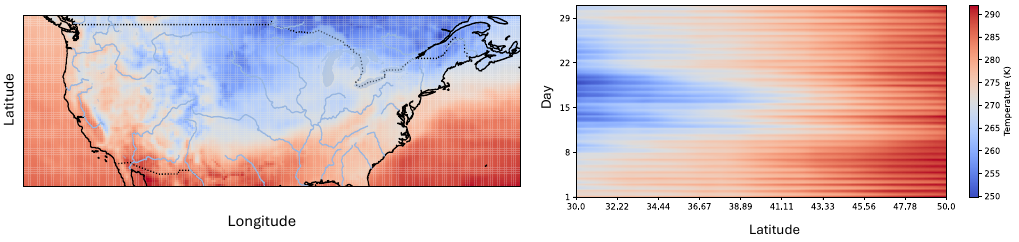}
    \caption{Left plot: Geographic distribution of 2-meter temperature at the beginning of January 2024. Right plot: Temporal evolution of 2-meter temperature averaged over longitude for January 2024.}
    \label{fig:app_era5}
\end{figure*}

\paragraph{Real world data manifold.} 
In addition to the synthetic static and dynamic datasets, we tested our framework on a real world ERA5-Land dataset \cite{munoz_sabater_2019}. Specifically, the data consists of 2-meter temperature measurements for January 2024, sampled at six-hour intervals (00:00, 04:00, 08:00, 12:00, 16:00, and 20:00 UTC). The geographic area covered ranges from 30°N to 50°N latitude and from 60°W to 130°W longitude. To simplify the analysis, the 2-meter temperature data was averaged over the longitude dimension to reduce the dimensionality. Overall this process resulted in dataset of 81 dimensions. A visualization of the initial 2-meter temperature data and the temperature data averaged over the longitude dimension for January 2024 using the ERA5-Land dataset is given in Fig. \ref{fig:app_era5}.

\paragraph{Implementation and Training.}
We construct a time-dependent proximity graph, where nodes represent time points with $d$-dimensional node attributes, which encode the spatial discretization. Connectivity is inferred via proximity, i.e., each node is connected to its $k$ nearest neighbors. We then utilize manifold learning to learn low-dimensional node representations.

\subsection{Comparative analysis of manually tuned manifold learning methods}
\begin{table*}[!h]
\centering
\begin{scriptsize}
\caption{Numerical results. Q-scores range from 0 (worst) to 1 (best). 
}
\begin{tabular}{c*{9}{c}} 
      \toprule
       & \multicolumn{3}{l}{\textbf{fKdV}} & \multicolumn{3}{l}{\textbf{KS}} & \multicolumn{3}{l}{\textbf{SG}} \\
      \cmidrule(r){2-4} \cmidrule(r){5-7}  \cmidrule{8-10} 
      {} & $Q_{local}$ & $Q_{global}$ & $K_{\max}$ & 
       $Q_{local}$ & $Q_{global}$ & $K_{\max}$ &  $Q_{local}$ & $Q_{global}$ & $K_{\max}$
      \tabularnewline
      \midrule
       \textsc{PCA} & 0.55 & 0.91 & 3 & \textbf{0.99} & \textbf{0.95} & 3 & \textbf{0.79} & 0.94 & 80 \\
       \textsc{MDS} & \textbf{0.75} & 0.95 & 214 & 0.96 & 0.94 & 6 & 0.74 & 0.93 & 19 \\
       \textsc{Isomap} & 0.63 & \textbf{0.97} & 247 & 0.84 & 0.94 & 1 & 0.77 & \textbf{0.95} & 71 \\ 
       \textsc{LLE} & 0.44 & 0.68 & 4 & 0.64 & 0.90 & 302 & 0.44 & 0.76 & 3 \\
       \textsc{SE} & 0.61 & 0.84 & 201 & 0.76 & 0.72 & 5 & 0.48 & 0.79 & 23\\
       \midrule
       \textsc{DW} & 0.52 (0.06) & 0.76 (0.06) & 156.1 (34.0) & 0.38 (0.02) & 0.57 (0.01) & 10.7 (0.9) & 0.39 (0.03) & 0.71 (0.03)  & 63.7 (11.2)\\ 
       \textsc{GCN} & 0.47 (0.06) & 0.73 (0.05) & 41.5 (56.2) & 0.50 (0.04) & 0.77 (0.07) & 40.4 (107.9) & 0.43 (0.03) & 0.72 (0.05) & 48.3 (16.3) \\
       \textsc{GS} & 0.37 (0.08) & 0.70 (0.07) & 91.9 (90.3) & 0.82 (0.02) & 0.81 (0.05) & 4.0 (0.0) & 0.35 (0.06) & 0.66 (0.05) & 30.0 (20.8) \\ 
       \textsc{GAT} & 0.43 (0.03) & 0.68 (0.03) & 17.2 (3.2) & 0.79 (0.01) & 0.84 (0.03) & 4.0 (0.0) & 0.45 (0.02) & 0.73 (0.04) & 19.8 (0.8) \\
       \bottomrule
    \end{tabular}
    \label{tab:results_table}
    \end{scriptsize}
  \end{table*}

  \begin{table*}[!h]
\centering
\begin{scriptsize}
    \caption{Comparison of optimization routines.}
\begin{tabular}{c*{9}{c}} 
      \toprule
       &  \multicolumn{3}{l}{\textbf{Bayesian}} & \multicolumn{3}{l}{\textbf{Random Search}} & \multicolumn{3}{l}{\textbf{Manual}}  \\
      \cmidrule(r){2-4} \cmidrule(r){5-7} \cmidrule(r){8-10} 
      {} & Model & Accuracy & Time (s) & Model & Accuracy & Time (s) & 
       Model & Accuracy & Time (s)
      \tabularnewline
      \midrule
       fKdV & PCA &  \makecell{$Q_{local}=0.86$\\ $Q_{global}=0.99$} & 71.06
       & PCA & \makecell{$Q_{local}=0.86$\\ $Q_{global}=0.99$} & 59.95  & MDS & \makecell{$Q_{local}=0.75$\\ $Q_{global}=0.97$} & {7292.30}\\
       KS & MDS & \makecell{$Q_{local}=0.96$\\ $Q_{global}=0.99$} & 112.83 & PCA & \makecell{$Q_{local}=0.99$\\ $Q_{global}=0.99$} & 87.21  & PCA, MDS & \makecell{$Q_{local}=0.99$\\ $Q_{global}=0.99$} & {7148.78}\\
       SG & PCA & \makecell{$Q_{local}=0.76$\\ $Q_{global}=0.96$} & 78.35 & PCA & \makecell{$Q_{local}=0.96$\\ $Q_{global}=0.99$} & 64.99  & PCA, MDS & \makecell{$Q_{local}=0.88$\\ $Q_{global}=0.99$} & {7897.56}\\
       \midrule 
       Swiss roll & MDS & \makecell{$Q_{local}=0.99$\\ $Q_{global}=1.0$} & 45.26 & Isomap & \makecell{$Q_{local}=0.87$\\ $Q_{global}=0.91$} & 36.01 & MDS & \makecell{$Q_{local}=0.99$ \\ $Q_{global}=0.99$} & 4061.08 \\
       Sphere & Isomap & \makecell{$Q_{local}=0.99$\\ $Q_{global}=1.0$} & 35.76 & MDS & \makecell{$Q_{local}=0.98$\\ $Q_{global}=0.99$} & 36.3 & MDS & \makecell{$Q_{local}=0.99$\\ $Q_{global}=1.0$} & 4067.18\\
       \midrule 
       ERA5 &  Isomap & \makecell{$Q_{local}=0.93$\\ $Q_{global}=0.99$} & 102.96 & PCA & \makecell{$Q_{local}=0.93$\\ $Q_{global}=1.0$} & 91.17 & PCA, MDS & \makecell{$Q_{local}=0.93$ \\ $Q_{global}=1.0$} & 1441.68 \\
       \bottomrule
    \end{tabular}
    \label{tab:results_auto}
    \end{scriptsize}
  \end{table*}
  
  In a first set of experiments, we run all approaches for all three  case studies with some manual hyperparameter tuning. We utilize the results for a comparative analysis of the learned representations and the ground truth (see Fig.~\ref{fig:app_fkdv}). We run the deep learning-based approaches for 1000 epochs using an SGD optimizer with learning rate $10^{-2}$ and a negative sampling loss. For \textsc{GCN, GraphSage} and \textsc{GAT} we use two hidden layers with 256 and 512 units. For both the classical and deep learning methods, we construct similarity graphs using $k=20$ nearest neighbours. For ease of visualization, we show results in three dimensions. Further implementation details and hyperparameter choices can be found in Apx.~\ref{sec:dl-params}.
  Our results indicate that the classical manifold learning approaches, specifically global methods (PCA, MDS, Isomap) perform best across our test data sets.
  
\paragraph{Qualitative and Quantitative Analysis.} A crucial aspect of any reduced-order dynamical model is to ensure the geometry of the higher dimensional data is faithfully conserved, i.e., is capturing the underlying dynamics of the system. To compare the classical and deep learning based methods, we apply all approaches to the KS bursting dynamics data and the travelling wave data from both the fKdV and SG equations. We compare the learned three-dimensional reduced order dynamics with ground truth representations obtained by a Fourier projection on the two dominant modes of the raw data (see Fig.~\ref{fig:results}); a comprehensive comparison of all methods across all three case studies can be found in Appendix~\ref{apx:exp}. We also quantitatively measure the quality of the reduced order dynamics using local and global Q-score metrics (Table \ref{tab:results_table}). For all the deep learning models we run 10 different initializations with different seed numbers giving the mean and standard deviation values, where the latter is the value within the parenthesis in Table \ref{tab:results_table}.


\paragraph{Characterizing local and global geometric features.}
Comparing the learned dynamics both qualitatively and quantitatively, we observe that \textsc{PCA} and \textsc{MDS} perform well with respect to local and global Q-scores, as well as with respect to visual inspection.  This may indicate that variation in the data can be captured well within linear subspaces. \textsc{Isomap}, the third global manifold learning approach tested here, performs competitively; the local methods (\textsc{LLE} and \textsc{SE}) perform low with respect to all evaluation criteria. This suggest that a successful approach needs to accurately capture both similarity between nearby points and dissimilarity between far distant points, which can only be achieved by global manifold learning approaches. The inability of random-walk based similarity measures to capture global pairwise distances may stem from approximation errors inherent to their probabilistic nature, but further analysis is needed to explain the performance trade-offs between classical and deep learning based methods.


\subsection{Benchmarking Automated Model Selection}
\label{sec-auto-experiment}
\noindent A robust and scalable automation framework should reduce the runtime while achieving comparable or superior accuracy than traditional manual model selection. To investigate this, we tracked the computation time of the automated pipeline against running the individual models for this experiment. To ensure a fair comparison of classical deep learning-based manifold learning methods, all algorithms were run using CPUs.
To analyze the utility of the automated pipeline we conduct a second set of experiments where we compare the quality of the learned representations obtained via automated and manual model selection. Results for all three physical systems can be found in Table~\ref{tab:results_auto}. We further include results for synthetic static data (swiss role, sphere) and a real dynamic data set (ERA5).

We observe that classical global methods, i.e., Isomap, PCA, and MDS, perform best in all settings, confirming our expectation from the first set of experiments. Automatically tuning hyperparameters generally achieves a higher accuracy in less time than manual tuning. The difference in runtime between the automated and manual model selection are striking: The runtime of manual model selection is two orders of magnitude higher than that of the automated pipeline. This result highlights the utility of our automated framework, whereby we have achieved a significant improvement to the quality of the learnt embedding in addition to a significant decrease in computation time. Additional details and experimental results can be found in Apx.~\ref{apx:model-select}. A comparison of our two automated approaches indicates that random search is the fastest, but that the Bayesian approach can sometimes achieve a higher accuracy with only slightly higher compute time. 


\section{Discussion}\label{sec:discussion}
   In this paper we investigated the utility of manifold learning approaches for recovering reduced-order dynamics in spatial-temporal data. Our systematic quantitative and qualitative study of representation trade-offs suggests that manifold learning is capable of learning dynamics, however, the quality of the learned representations varies highly across model architectures and hyperparameter choices. Manually fine-tuning these design choices can be prohibitively expensive in practise.  
Motivated by this observation, we propose an \emph{Automated Manifold Learning} framework, which performs automated model selection and hyperparameter tuning. Our experimental results suggest that this framework improves the representation quality and scalability in reduced order modeling substantially, reducing the runtime by orders of magnitudes in comparison with manual model selection.

While AutoML frameworks have been applied to a variety of learning and inference tasks, we believe that our framework is the first to integrate automated model selection into manifold learning. Our experimental results present a proof of concept that demonstrates improved scalability and representation quality, while retaining the transparency and interpretability of manifold learning for reduced-order modeling. While our work focuses on reduced-order modeling, automated manifold learning can be applied in a wide range of applications, including shape analysis, visualization of scientific data and computer graphics, and should therefore be of independent interest. In future work, we aim to extend our study to a broader range of case studies and representation learning methods with the aim of understanding trade-offs and commonalities between these approaches more comprehensively. 

A limitation of the present paper is the relatively small set of quality metrics utilized in the automation pipeline. While the results derived from Q-scores align with our observations from a visual inspection of the learned data, we believe that an expanded set of quality metrics could improve the automated model selection. In particular, in future work we hope to include representation quality metrics that evaluate the preservation of local and global geometric features more explicitly, as well as metrics that assess the utility of the learned representations in downstream tasks. We further aim to investigate the effects of non-uniform sampling to gain better insight into the applicability of the tested approaches to real-world data. We believe that there are opportunities for optimizing the model selection procedure, including via an integration of hyperparameter heuristics or warm-starts. Finally, we hope to integrate the proposed pipeline into existing manifold learning libraries.


\bibliography{bibliography}

\begin{thebibliography}{42}
\providecommand{\natexlab}[1]{#1}
\providecommand{\url}[1]{\texttt{#1}}
\expandafter\ifx\csname urlstyle\endcsname\relax
  \providecommand{\doi}[1]{doi: #1}\else
  \providecommand{\doi}{doi: \begingroup \urlstyle{rm}\Url}\fi

\bibitem[Bakarji et~al.(2023)Bakarji, Champion, Nathan~Kutz, and Brunton]{bakarji2023_discovering}
Joseph Bakarji, Kathleen Champion, J~Nathan~Kutz, and Steven~L Brunton.
\newblock Discovering governing equations from partial measurements with deep delay autoencoders.
\newblock \emph{Proceedings of the Royal Society A}, 479\penalty0 (2276):\penalty0 20230422, 2023.

\bibitem[Belkin and Niyogi(2003)]{belkin2003laplacian}
Mikhail Belkin and Partha Niyogi.
\newblock Laplacian eigenmaps for dimensionality reduction and data representation.
\newblock \emph{Neural computation}, 15\penalty0 (6):\penalty0 1373--1396, 2003.

\bibitem[Bergstra and Bengio(2012)]{bergstra12a}
James Bergstra and Yoshua Bengio.
\newblock Random search for hyper-parameter optimization.
\newblock \emph{Journal of Machine Learning Research}, 13\penalty0 (10):\penalty0 281--305, 2012.

\bibitem[Binder(2019)]{fkdv_2019}
Benjamin~J. Binder.
\newblock Steady two-dimensional free-surface flow past disturbances in an open channel: Solutions of the korteweg–de vries equation and analysis of the weakly nonlinear phase space.
\newblock \emph{Fluids}, 4\penalty0 (1), 2019.
\newblock ISSN 2311-5521.
\newblock \doi{10.3390/fluids4010024}.
\newblock URL \url{https://www.mdpi.com/2311-5521/4/1/24}.

\bibitem[Binder and Vanden-Broeck(2005)]{binder2005_fkdv}
BJ~Binder and J-M Vanden-Broeck.
\newblock Free surface flows past surfboards and sluice gates.
\newblock \emph{European Journal of Applied Mathematics}, 16\penalty0 (5):\penalty0 601--619, 2005.

\bibitem[Bronstein et~al.(2006)Bronstein, Bronstein, and Kimmel]{bronstein2006generalized}
Alexander~M Bronstein, Michael~M Bronstein, and Ron Kimmel.
\newblock Generalized multidimensional scaling: a framework for isometry-invariant partial surface matching.
\newblock \emph{Proceedings of the National Academy of Sciences}, 103\penalty0 (5):\penalty0 1168--1172, 2006.

\bibitem[Brunton et~al.(2016)Brunton, Proctor, and Kutz]{brunton2016discovering}
Steven~L Brunton, Joshua~L Proctor, and J~Nathan Kutz.
\newblock Discovering governing equations from data by sparse identification of nonlinear dynamical systems.
\newblock \emph{Proceedings of the national academy of sciences}, 113\penalty0 (15):\penalty0 3932--3937, 2016.

\bibitem[Champion et~al.(2019)Champion, Lusch, Kutz, and Brunton]{champion2019_partialdata}
Kathleen Champion, Bethany Lusch, J~Nathan Kutz, and Steven~L Brunton.
\newblock Data-driven discovery of coordinates and governing equations.
\newblock \emph{Proceedings of the National Academy of Sciences}, 116\penalty0 (45):\penalty0 22445--22451, 2019.

\bibitem[Chazal and Michel(2021)]{tda}
Frédéric Chazal and Bertrand Michel.
\newblock An introduction to topological data analysis: Fundamental and practical aspects for data scientists.
\newblock \emph{Frontiers in Artificial Intelligence}, 4, 2021.

\bibitem[Fasina et~al.(2022)Fasina, Krishnaswamy, and Krishnapriyan]{fasina_geometric_nodate}
Oluwadamilola Fasina, Smita Krishnaswamy, and Aditi Krishnapriyan.
\newblock Geometric {NeuralPDE} ({GNPnet}) {Models} for {Learning} {Dynamics}.
\newblock 2022.

\bibitem[Fung et~al.(2016)Fung, Hanna, Vendrell, Ramakrishna, Seideman, Santra, and Ourmazd]{fung2016_dynamics}
Russell Fung, Ataya~M Hanna, Oriol Vendrell, S~Ramakrishna, Tamar Seideman, Robin Santra, and Abbas Ourmazd.
\newblock Dynamics from noisy data with extreme timing uncertainty.
\newblock \emph{Nature}, 532\penalty0 (7600):\penalty0 471--475, 2016.

\bibitem[Grover and Leskovec(2016)]{grover2016node2vec}
Aditya Grover and Jure Leskovec.
\newblock node2vec: Scalable feature learning for networks.
\newblock In \emph{Proceedings of the 22nd ACM SIGKDD international conference on Knowledge discovery and data mining}, pages 855--864, 2016.

\bibitem[Hamilton et~al.(2017)Hamilton, Ying, and Leskovec]{hamilton}
Will Hamilton, Zhitao Ying, and Jure Leskovec.
\newblock Inductive representation learning on large graphs.
\newblock \emph{Advances in neural information processing systems}, 30, 2017.

\bibitem[Hirota(1972)]{hirota1972_SG}
Ryogo Hirota.
\newblock Exact solution of the sine-gordon equation for multiple collisions of solitons.
\newblock \emph{Journal of the Physical Society of Japan}, 33\penalty0 (5):\penalty0 1459--1463, 1972.

\bibitem[Hutter et~al.(2011)Hutter, Hoos, and Leyton-Brown]{hutter2011smbo}
Frank Hutter, Holger~H Hoos, and Kevin Leyton-Brown.
\newblock Sequential model-based optimization for general algorithm configuration.
\newblock In \emph{Learning and Intelligent Optimization: 5th International Conference, LION 5, Rome, Italy, January 17-21, 2011. Selected Papers 5}, pages 507--523. Springer, 2011.

\bibitem[Kipf and Welling(2017)]{Kipf_gcn_2017}
Thomas~N. Kipf and Max Welling.
\newblock {Semi-Supervised Classification with Graph Convolutional Networks}.
\newblock In \emph{Proceedings of the 5th International Conference on Learning Representations}, 2017.

\bibitem[Kirby and Armbruster(1992)]{kirby1992_ks_bursting}
Michael Kirby and Dieter Armbruster.
\newblock Reconstructing phase space from pde simulations.
\newblock \emph{Zeitschrift f{\"u}r angewandte Mathematik und Physik ZAMP}, 43:\penalty0 999--1022, 1992.

\bibitem[Kruskal(1964)]{kruskal1964multidimensional}
Joseph~B Kruskal.
\newblock Multidimensional scaling by optimizing goodness of fit to a nonmetric hypothesis.
\newblock \emph{Psychometrika}, 29\penalty0 (1):\penalty0 1--27, 1964.

\bibitem[Kuramoto(1978)]{kuramoto_1978}
Yoshiki Kuramoto.
\newblock {Diffusion-Induced Chaos in Reaction Systems}.
\newblock \emph{Progress of Theoretical Physics Supplement}, 64:\penalty0 346--367, 02 1978.
\newblock ISSN 0375-9687.
\newblock \doi{10.1143/PTPS.64.346}.
\newblock URL \url{https://doi.org/10.1143/PTPS.64.346}.

\bibitem[Lee and Verleysen(2010)]{lee_scale-independent_2010}
John~A. Lee and Michel Verleysen.
\newblock Scale-independent quality criteria for dimensionality reduction.
\newblock \emph{Pattern Recognition Letters}, 31\penalty0 (14):\penalty0 2248--2257, October 2010.

\bibitem[Lubold et~al.(2023)Lubold, Chandrasekhar, and McCormick]{lubold2023identifying}
Shane Lubold, Arun~G Chandrasekhar, and Tyler~H McCormick.
\newblock Identifying the latent space geometry of network models through analysis of curvature.
\newblock \emph{Journal of the Royal Statistical Society Series B: Statistical Methodology}, 85\penalty0 (2):\penalty0 240--292, 2023.

\bibitem[Lusch et~al.(2018)Lusch, Kutz, and Brunton]{lusch2018deep}
Bethany Lusch, J~Nathan Kutz, and Steven~L Brunton.
\newblock Deep learning for universal linear embeddings of nonlinear dynamics.
\newblock \emph{Nature communications}, 9\penalty0 (1):\penalty0 4950, 2018.

\bibitem[Muñoz~Sabater(2019)]{munoz_sabater_2019}
J.~Muñoz~Sabater.
\newblock Era5-land hourly data from 1950 to present, 2019.
\newblock URL \url{https://doi.org/10.24381/cds.e2161bac}.
\newblock Accessed on 26-Jul-2024.

\bibitem[Nasim and de~Sousa~Almeida(2024)]{nasim_JLS_scml}
Imran Nasim and Joao~Lucas de~Sousa~Almeida.
\newblock Using neural implicit flow to represent latent dynamics of canonical systems.
\newblock In \emph{International Conference on Scientific Computing and Machine Learning}, 2024.

\bibitem[Nasim and Henderson(2024)]{nasim2024_dynamicalAE}
Imran Nasim and Michael~E Henderson.
\newblock Dynamically meaningful latent representations of dynamical systems.
\newblock \emph{Mathematics}, 12\penalty0 (3):\penalty0 476, 2024.

\bibitem[Nasim and Weber(2024)]{scml}
Imran Nasim and Melanie Weber.
\newblock Learning reduced order dynamics via geometric representations.
\newblock In \emph{International Conference on Scientific Computing and Machine Learning}, 2024.

\bibitem[Otto and Rowley(2019)]{otto_linearly_2019}
Samuel~E. Otto and Clarence~W. Rowley.
\newblock Linearly {Recurrent} {Autoencoder} {Networks} for {Learning} {Dynamics}.
\newblock \emph{SIAM Journal on Applied Dynamical Systems}, 18, January 2019.

\bibitem[Pedregosa et~al.(2011)Pedregosa, Varoquaux, Gramfort, Michel, Thirion, Grisel, Blondel, Prettenhofer, Weiss, Dubourg, Vanderplas, Passos, Cournapeau, Brucher, Perrot, and Duchesnay]{scikit-learn}
F.~Pedregosa, G.~Varoquaux, A.~Gramfort, V.~Michel, B.~Thirion, O.~Grisel, M.~Blondel, P.~Prettenhofer, R.~Weiss, V.~Dubourg, J.~Vanderplas, A.~Passos, D.~Cournapeau, M.~Brucher, M.~Perrot, and E.~Duchesnay.
\newblock Scikit-learn: Machine learning in {P}ython.
\newblock \emph{Journal of Machine Learning Research}, 12:\penalty0 2825--2830, 2011.

\bibitem[Perozzi et~al.(2014)Perozzi, Al-Rfou, and Skiena]{Deepwalk_perozzi_2014}
Bryan Perozzi, Rami Al-Rfou, and Steven Skiena.
\newblock Deepwalk: Online learning of social representations.
\newblock In \emph{Proceedings of the 20th ACM SIGKDD International Conference on Knowledge Discovery and Data Mining}, page 701–710. Association for Computing Machinery, 2014.
\newblock ISBN 9781450329569.

\bibitem[Roweis and Saul(2000)]{roweis2000nonlinear}
Sam~T Roweis and Lawrence~K Saul.
\newblock Nonlinear dimensionality reduction by locally linear embedding.
\newblock \emph{science}, 290\penalty0 (5500):\penalty0 2323--2326, 2000.

\bibitem[Rozemberczki et~al.(2020)Rozemberczki, Kiss, and Sarkar]{rozemberczki2020_karateclub}
Benedek Rozemberczki, Oliver Kiss, and Rik Sarkar.
\newblock {Karate Club: An API Oriented Open-source Python Framework for Unsupervised Learning on Graphs}.
\newblock In \emph{Proceedings of the 29th ACM International Conference on Information and Knowledge Management (CIKM '20)}, page 3125–3132. ACM, 2020.

\bibitem[Shi and Malik(2000)]{shi2000_SE}
Jianbo Shi and Jitendra Malik.
\newblock Normalized cuts and image segmentation.
\newblock \emph{IEEE Transactions on pattern analysis and machine intelligence}, 22\penalty0 (8):\penalty0 888--905, 2000.

\bibitem[Snoek et~al.(2012)Snoek, Larochelle, and Adams]{snoek2012_bayesian}
Jasper Snoek, Hugo Larochelle, and Ryan~P Adams.
\newblock Practical bayesian optimization of machine learning algorithms.
\newblock \emph{Advances in neural information processing systems}, 25, 2012.

\bibitem[Tenenbaum et~al.(2000)Tenenbaum, Silva, and Langford]{tenenbaum2000global}
Joshua~B Tenenbaum, Vin~de Silva, and John~C Langford.
\newblock A global geometric framework for nonlinear dimensionality reduction.
\newblock \emph{science}, 290\penalty0 (5500):\penalty0 2319--2323, 2000.

\bibitem[Tu et~al.(2019)Tu, Cui, Wang, Wang, and Zhu]{tu2019AutoNE}
Ke~Tu, Peng Cui, Xiao Wang, Fei Wang, and Wenwu Zhu.
\newblock Autone: Hyperparameter optimization for massive network embedding.
\newblock In \emph{Proceedings of the 25th ACM SIGKDD International Conference on Knowledge Discovery \& Data Mining}. ACM, 2019.

\bibitem[Velickovic et~al.(2017)Velickovic, Cucurull, Casanova, Romero, Lio, Bengio, et~al.]{Velickovic_GATS_2017}
Petar Velickovic, Guillem Cucurull, Arantxa Casanova, Adriana Romero, Pietro Lio, Yoshua Bengio, et~al.
\newblock Graph attention networks.
\newblock \emph{stat}, 1050\penalty0 (20):\penalty0 10--48550, 2017.

\bibitem[Wang et~al.(2019)Wang, Sun, Liu, Sarma, Bronstein, and Solomon]{wang2019dynamic}
Yue Wang, Yongbin Sun, Ziwei Liu, Sanjay~E Sarma, Michael~M Bronstein, and Justin~M Solomon.
\newblock Dynamic graph cnn for learning on point clouds.
\newblock \emph{ACM Transactions on Graphics (tog)}, 38\penalty0 (5):\penalty0 1--12, 2019.

\bibitem[Weber(2020)]{weber2020neighborhood}
Melanie Weber.
\newblock Neighborhood growth determines geometric priors for relational representation learning.
\newblock In \emph{International Conference on Artificial Intelligence and Statistics}, volume 108, pages 266--276, 2020.

\bibitem[Whalen et~al.(2015)Whalen, Brio, and Moloney]{whalen2015_etd}
Patrick Whalen, Moysey Brio, and Jerome~V Moloney.
\newblock Exponential time-differencing with embedded runge--kutta adaptive step control.
\newblock \emph{Journal of Computational Physics}, 280:\penalty0 579--601, 2015.

\bibitem[Yang et~al.(2020)Yang, Fan, Wu, and Udell]{yang2020efficient}
Chengrun Yang, Jicong Fan, Ziyang Wu, and Madeleine Udell.
\newblock Efficient automl pipeline search with matrix and tensor factorization.
\newblock \emph{arXiv preprint arXiv:2006.04216}, 2020.

\bibitem[Yang et~al.(2022)Yang, Bender, Liu, Kindermans, Udell, Lu, Le, and Huang]{yang2022tabnas}
Chengrun Yang, Gabriel Bender, Hanxiao Liu, Pieter-Jan Kindermans, Madeleine Udell, Yifeng Lu, Quoc~V Le, and Da~Huang.
\newblock Tabnas: Rejection sampling for neural architecture search on tabular datasets.
\newblock \emph{Advances in Neural Information Processing Systems}, 35:\penalty0 11906--11917, 2022.

\bibitem[Zhang et~al.(2021)Zhang, Wang, and Zhu]{zhang2021_automlgraph_survey}
Ziwei Zhang, Xin Wang, and Wenwu Zhu.
\newblock Automated machine learning on graphs: A survey.
\newblock \emph{arXiv preprint arXiv:2103.00742}, 2021.

\end{thebibliography}

\newpage
\onecolumn
\appendix
\newpage
\section{Methods for Geometric Representation Learning}\label{sec:app-methods}
\begin{table}[!ht]
\small
\centering
\caption{Overview of Geometric Representation Learning methods.}
\begin{tabular}{ccccc}
      \toprule
      \textbf{Algorithm} & \makecell{Type} & \makecell{Geometry preserved}  & Method & Embedding Dimension\\
      \midrule
      \textsc{PCA} & spectral & \makecell{global proximity} & \makecell{top eigenvalues}  & learnt \\
     \textsc{MDS} & spectral & \makecell{global isometry}   & \makecell{top eigenvalues} & learnt  \\
      \textsc{Isomap} & spectral & \makecell{global isometry} & top eigenvalues & learnt  \\
     \textsc{LLE} & spectral & \makecell{conformal, local poximity} & bottom eigenvalues & hyperparameter  \\
    \textsc{SE} & spectral & local proximity  & bottom eigenvalues & hyperparameter  \\
     \textsc{DeepWalk} & shallow &  local proximity &  \makecell{probabilistic} & hyperparameter\\
     \textsc{GCN} & deep & local proximity & \makecell{probabilistic} & hyperparameter \\
     \textsc{GraphSage} & deep & local proximity & \makecell{probabilistic} & hyperparameter  \\
     \textsc{GAT} & deep & local proximity & probabilistic & hyperparameter  \\
      \bottomrule
    \end{tabular}
    \label{tab:mf-taxonomy}
\end{table}
The methods considered in this study identify geometric structure in a purely data-driven manner, without prior characterization of the data's geometry. We note that there is a complementary body of literature, which provides tools for characterizing data geometry intrinsically, independent of learned data representations~\citep{tda,lubold2023identifying,weber2020neighborhood}.

\subsection{Details on Classical Manifold Learning algorithms}
\paragraph{Multi-Dimensional Scaling (MDS).}
Multi-dimensional scaling is a popular method for Representation Learning in both Euclidean (cMDS~\citep{kruskal1964multidimensional}) and non-Euclidean spaces (gMDS~\citep{bronstein2006generalized}).  It computes a map $\phi$ by directly minimizing the representation error, i.e., the difference between pairwise distances $d_\Xc(x,y)$ in the data space and the distances of the corresponding manifold representations $d_{\widetilde{\Mc}}(\phi(x),\phi(y))$ over all pairs $x,y \in \Xc$. 

\paragraph{Isometric Feature Mapping (Isomap).} Isomap~\citep{tenenbaum2000global} is an extension of of MDS, which approximates geodesic distances via shortest paths on the constructed proximity graph and then performs an MDS step.

\paragraph{Locally Linear Embeddings (LLE).}
The focus of local manifold learning methods,  such as \emph{Locally Linear Embeddings} (LLE)~\citep{roweis2000nonlinear}, is on preserving local similarity structure.  A local similarity matrix is constructed with respect to the $k$ nearest neighbors of a data point from which local structure is extracted using spectral decomposition. 

\paragraph{Spectral Embeddings (SE).} 
SE~\cite{shi2000_SE} is a spectral clustering technique used for non-linear manifold learning. It typically uses Laplacian Eigenmaps, utilizing the spectral properties of the Laplacian matrix of the constructed graph.

\subsection{Details on Deep Learning-based Manifold Learning algorithms}
\paragraph{DeepWalk.}
DeepWalk~\citep{Deepwalk_perozzi_2014} is an early approach for shallow graph embeddings. It learns a latent space representation of nodes based utilizing a random walk based similarity measure. 

\paragraph{Graph Convolutional Networks and GraphSage.}
One of the early and still widely popular GNN architectures are
GCN~\citep{Kipf_gcn_2017} and GraphSage~\citep{hamilton}, both of which extend the idea of convolutional neural networks to the graph domain. Previous applications of graph convolutions to point clouds were considered in~\citep{wang2019dynamic}.

\paragraph{Graph Attention.}
Graph attention, which translates the idea of \emph{transformers} to Graph Neural Networks, was first proposed by~\citep{Velickovic_GATS_2017}. The most popular variants of Graph Attention Networks are based on this architecture.

\section{Implementation of Manifold Learning algorithms}
\label{sec:dl-params}

\paragraph{Classical methods.} All of the classical models considered in this study (\textsc{LLE, Isomap, Spectral Embedding, MDS, PCA})
were implemented using the \textsc{scikit-learn} python library \cite{scikit-learn}. The proximity graph was constructed using a $k$ nearest neighbours approach setting $k=20$.

\paragraph{Deep Learning based methods.}
For the results in Table \ref{tab:results_table} we list the Parameters used in \textsc{GCN}, \textsc{Graphsage} and \textsc{GAT} methods. The number of neurons in the hidden layers were 256 and 512 where each layer used the ReLU activation function. We used the SGD optimizer with a learning rate of $10^{-2}$ and ran all deep learning based models for 1000 epochs.

For \textsc{GCN, GraphSage} and \textsc{GAT} we choose a fixed walk length of 5 and use a single negative sample. For \textsc{GAT} we use a single attention head.
Through our experiments we found that the SGD optimizer yielded more accurate results compared to Adam and found the commonly used learning rate of $10^{-2}$ to work well for our applications. The \textsc{DeepWalk} method was implemented via \textsc{Karateclub} package \citep{rozemberczki2020_karateclub} which we ran for 1000 epochs with a walk length of 50 and a window size of 5. We found that increasing the walk length yielded a better metric score but was more computationally expensive. We did not observe a significant increase in the performance for walk lengths greater than 50.

\section{Details on Automated Model Selection and Comparison Experiment}
\label{apx:model-select}

\begin{table}[h!]
    \centering
    \caption{Hyperparameter definitions and values used for the Automated Model Selection framework for the dynamic datasets and the ERA climate dataset.}
    \begin{tabular}{|m{5cm}|m{10cm}|}
        \hline
        \textbf{Parameter} & \textbf{Values} \\
        \hline
        Optimizer's learning rate ($l_r$) & Linearly spaced within the range $[10^{-4}, 10^{-2}]$ (5 values) \\
        \hline
        Optimization algorithm & Adam, SGD \\
        \hline
        Number of hidden neurons ($h_n$) & \{128, 256, 512, 1028\} \\
        \hline
        Dimension of the output features ($n_{dim}$) & \{3, 4, 5\} \\
        \hline
        Number of nearest neighbors ($k$) & \{40, 50, 60, 70\} \\
        \hline
        Length of random walks ($w$) & \{50, 60, 70\} \\
        \hline
        Walk length for constructing subgraphs ($w_{sub}$) & \{80, 90, 100\} \\
        \hline
    \end{tabular}
    \label{tab:hyperparameters_dynamical}
\end{table}

\begin{table}[h!]
    \centering
    \caption{Hyperparameter definitions and values used for the Automated Model Selection framework for the static datasets (Sphere and Swiss roll).}
    \begin{tabular}{|m{5cm}|m{10cm}|}
        \hline
        \textbf{Parameter} & \textbf{Values} \\
        \hline
        Optimizer's learning rate ($l_r$) & Linearly spaced within the range $[10^{-4}, 10^{-2}]$ (5 values) \\
        \hline
        Optimization algorithm & Adam, SGD \\
        \hline
        Number of hidden neurons ($h_n$) & \{64, 128, 256, 512\} \\
        \hline
        Dimension of the output features ($n_{dim}$) & \{1, 2, 3\} \\
        \hline
        Number of nearest neighbors ($k$) & \{40, 50, 60, 70\} \\
        \hline
        Length of random walks ($w$) & \{20, 30, 40\} \\
        \hline
        Walk length for constructing subgraphs ($w_{sub}$) & \{60, 70, 80\} \\
        \hline
    \end{tabular}
    \label{tab:hyperparameters_static}
\end{table}

\paragraph{Automated manifold learning.}
In our automated pipeline, we optimized the following hyperparameters: the optimizer's learning rate ($l_r$), the optimization algorithm, the number of hidden neurons ($h_n$), the dimension of the output features ($n_{\text{dim}}$), the number of nearest neighbors ($k$) for constructing the main graph, the length of random walks ($w$) for sampling nodes in deep learning methods, and the walk length for constructing subgraphs ($w_{\text{sub}}$). For the specific range of hyperparameter values used in the automated pipeline for i) the dynamical and climate datasets, see Table \ref{tab:hyperparameters_dynamical} ii) static datasets, see Table \ref{tab:hyperparameters_static}. When optimizing the \emph{Q-scores} for the deep learning based models on the subgraphs, we run all models for 100 epochs. Whereas for the main graph with the final optimized hyperparameters, we use 1000 epochs which matches the number of epochs used in our standard non-optimized framework.
For the computation time of the experiment described in Section \ref{sec-auto-experiment}, all models were run using in a single CPU core of an Apple M1 Max chip. The computation time of the automation framework was 28.36 seconds and the separate models took 278.44 seconds. The detailed breakdown of each model computation time in seconds is: 0.92 (\textsc{PCA}), 0.97 (\textsc{LLE}), 1.38 (\textsc{Isomap}), 3.67 (\textsc{MDS}), 0.92 (\textsc{SE}), 73.57 (\textsc{GAT}), 19.71 (\textsc{GCN}), 27.76 (\textsc{GraphSage}), 149.54 (\textsc{DeepWalk}). 

\begin{table}[ht!]
\centering
\caption{Numerical results running models individually for the fKdV dynamical data. $\lambda_t$ and $T_t$ correspond to the average time and total time to run the models in seconds. The standard deviations are displayed within parenthesis.  }
\begin{tabular}{l|ccccc}
\hline
       & $Q_{local}$ & $Q_{global}$ & $K_{\max}$ &  $\lambda_t$ & $T_t$  \\ \hline
\textsc{PCA}    & 0.58 (9.2e-2)  & 0.93 (2.0e-2)  & 3.0 (0) & \textbf{0.56} & \textbf{6.38}          \\
\textsc{MDS}    & \textbf{0.75} (6.0e-4)  & \textbf{0.97} (9.3e-3)  & 157.0 (51.3) & 5.83 & 59.14        \\
\textsc{LLE}    & 0.42 (4.0e-2)  & 0.70 (2.2e-2)  & 4.3 (2.9) & 0.82 & 8.96         \\
\textsc{Isomap} & 0.62 (2.6e-2)  & 0.95 (1.4e-3)  & 190.5 (25.4) & 1.38 & 14.60         \\
\textsc{SE}     & 0.61 (9.1e-3)  & 0.84 (1.7e-2)  & 197.6 (2.5) & 0.63 & 7.07         \\
\textsc{GCN}    & 0.43 (0.12)  & 0.69 (8.1e-2)  & 33.3 (55.8) & 51.54 & 516.12          \\
\textsc{GS}     & 0.39 (0.12)  & 0.68 (8.0e-2)  & 37.7 (48.9) & 169.02 & 1690.97        \\
\textsc{GAT}    &  0.46 (0.01)  & 0.68 (6.5e-2)  & 57.3 (61.7) & 324.70 & 3247.76         \\
\textsc{DeepWalk}   & 0.65 (2.2e-2)  & 0.88 (3.0e-2)  & 164.3 (53.7) & 174.05 & 1741.31        \\ \hline
\end{tabular}
\label{table:fkdv_experiment_individual}
\end{table}

\begin{table}[ht!]
\centering
\caption{Numerical results running models individually for the KS dynamical data. $\lambda_t$ and $T_t$ correspond to the average time and total time to run the models in seconds. The standard deviations are displayed within parenthesis. }
\begin{tabular}{l|ccccc}
\hline
       & $Q_{local}$ & $Q_{global}$ & $K_{\max}$ &  $\lambda_t$ & $T_t$  \\ \hline
\textsc{PCA}    & \textbf{0.99} (3.0e-3)  & \textbf{0.99} (1.0e-3)  & 2.7 (0.9) & \textbf{0.54}  & \textbf{6.21}          \\
\textsc{MDS}    & 0.97 (1.3e-4)  & 0.99 (1.8e-3)  & 4.6 (1.8) & 12.38 & 124.62        \\
\textsc{LLE}    & 0.67 (3.4e-3)  & 0.85 (1.4e-2)  & 151.3 (15.1) & 0.73 & 8.07         \\
\textsc{Isomap} & 0.93 (1.2e-2)  & 0.95 (4.0e-3)  & 3.9 (0.7) & 1.10  & 11.78         \\
\textsc{SE}     & 0.65 (3.4e-2)  & 0.72 (1.1e-2)  & 6.0 (2.1) & 0.62 & 6.94         \\
\textsc{GCN}    & 0.55 (5.9e-2)  & 0.61 (2.0e-2)  & 5.9 (2.7) & 53.35 & 534.24          \\
\textsc{GS}     & 0.58 (0.12)  & 0.68 (8.2e-2)  & 3.2 (1.6) & 165.57 & 1656.47        \\
\textsc{GAT}    &  0.58 (2.8e-2)  & 0.68 (1.0e-2)  & 40.9 (107.0) & 319.85  & 3199.25         \\
\textsc{DeepWalk}   & 0.44 (3.4e-3)  & 0.63 (1.6e-2)  & 33.1 (3.7) & 160.05 & 1601.20        \\ \hline
\end{tabular}
\label{table:ks_experiment_individual}
\end{table}

\begin{table}[ht!]
\centering
\caption{Numerical results running models individually for the SG dynamical data. $\lambda_t$ and $T_t$ correspond to the average time and total time to run the models in seconds. The standard deviations are displayed within parenthesis. }
\begin{tabular}{l|ccccc}
\hline
       & $Q_{local}$ & $Q_{global}$ & $K_{\max}$ &  $\lambda_t$ & $T_t$  \\ \hline
\textsc{PCA}    & \textbf{0.88} (2.7e-2)  & \textbf{0.99} (1.4e-2)  & 44.2 (14.0) & \textbf{0.57} & \textbf{6.55}          \\
\textsc{MDS}    & 0.86 (3.7e-3)  & 0.99 (3.3e-6)  & 47.9 (0.3) & 9.06 & 91.45        \\
\textsc{LLE}    & 0.54 (2.7e-2)  & 0.76 (1.2e-2) & 29.5 (14.3) & 0.77 & 8.54         \\
\textsc{Isomap} & 0.76 (1.2e-2)  & 0.96 (3.2e-3)  & 43.6 (20.9) & 1.35 & 14.29         \\
\textsc{SE}     & 0.63 (8.6e-3)  & 0.82 (6.7e-3)  & 189.4 (45.7) & 0.65 & 7.29         \\
\textsc{GCN}    & 0.37 (6.5e-2)  & 0.67 (7.6e-2)  & 49.9 (58.1) & 53.13 & 533.04          \\
\textsc{GS}     & 0.47 (0.11)  & 0.69 (8.2e-2)  & 44.1 (7.62) & 167.89 & 1679.65        \\
\textsc{GAT}    &  0.50 (7.3e-2)  & 0.72 (7.2e-2)  & 41.4 (6.4) & 324.94  & 3250.21         \\
\textsc{DeepWalk}   & 0.59 (6.9e-2)  & 0.73 (4.0e-2)  & 42.9 (6.4) & 230.59 & 2306.54        \\ \hline
\end{tabular}
\label{table:sg_experiment_individual}
\end{table}

\begin{table}[ht!]
\centering
\caption{Numerical results running models individually for the Swiss Roll static data $\lambda_t$ and $T_t$ correspond to the average time and total time to run the models in seconds. The standard deviations are displayed within parenthesis. }
\begin{tabular}{l|ccccc}
\hline
       & $Q_{local}$ & $Q_{global}$ & $K_{\max}$ &  $\lambda_t$ & $T_t$  \\ \hline
\textsc{PCA}    & \textbf{0.43} (1.9e-1)  & \textbf{0.80} (6.6e-2)  & 251.1 (83.7) & \textbf{0.59} & \textbf{6.69}          \\
\textsc{MDS}    & 0.37 (0.2)  & 0.74 (8.6e-2)  & 79.2 (26.4) & 4.8 & 48.85        \\
\textsc{LLE}    & 0.37 (0.2)  & 0.74 (7.6e-2) & 160.1 (86.8) & 0.81 & 8.88         \\
\textsc{Isomap} & 0.41 (1.7e-1)  & 0.79 (4.9e-2)  & 238.3 (76.9) & 1.55 & 16.24         \\
\textsc{SE}     & 0.33 (0.13)  & 0.72 (6.0e-2)  & 147.1 (56.8) & 0.63  & 7.10         \\
\textsc{GCN}    & 0.20 (9.9e-2)  & 0.60 (3.6e-2)  & 86.1 (26.9) & 34.10 & 341.82          \\
\textsc{GS}     & 0.25 (3.9e-2)  & 0.63 (1.7e-2)  & 82.1 (23.91) & 88.65 & 887.21        \\
\textsc{GAT}    &  0.27 (8.8e-2)  & 0.63 (2.8e-2)  & 86.6 (32.5) & 169.40 & 1694.78         \\
\textsc{DeepWalk}   & 0.24 (1.8e-2)  & 0.62 (1.2e-2)  & 81.9 (15.1) & 104.86 & 1049.51        \\ \hline
\end{tabular}
\label{table:swiss_roll_experiment_individual}
\end{table}

\begin{table}[ht!]
\centering
\caption{Numerical results running models individually for the Sphere static data $\lambda_t$ and $T_t$ correspond to the average time and total time to run the models in seconds. The standard deviations are displayed within parenthesis. }
\begin{tabular}{l|ccccc}
\hline
       & $Q_{local}$ & $Q_{global}$ & $K_{\max}$ &  $\lambda_t$ & $T_t$  \\ \hline
\textsc{PCA}  & {0.31} (2.1e-1)  & {0.73} (8.0e-2)  & 208.0 (62.4) & \textbf{0.57} & \textbf{6.45}          \\
\textsc{MDS}    & \textbf{0.45} (1.8e-1)  & \textbf{0.78} (7.5e-2)  & 138.0 (45.0) & 5.5 & 55.87        \\
\textsc{LLE}    & 0.36 (1.9e-1)  & 0.75 (7.5e-2) & 183.4 (54.1) & 0.80  & 8.74         \\
\textsc{Isomap} & 0.37 (0.2)  & 0.76 (7.8e-2) & 189.0 (62.4) & 1.56 & 16.44         \\
\textsc{SE}  & 0.36 (1.7e-1)  & 0.75 (6.8e-2) & 181.7 (44.3) & 0.63 & 7.10         \\
\textsc{GCN}    & 0.20 (9.9e-2)  & 0.60 (3.6e-2)  & 86.1 (26.9) & 34.10  & 341.82          \\
\textsc{GS}     & 0.20 (6.6e-2)  & 0.61 (3.7e-2)  & 95.2 (39.8) & 88.54 & 886.09        \\
\textsc{GAT}    &  0.24 (5.3e-2)  & 0.65 (5.2e-2)  & 125.6 (58.25) & 169.40 & 1694.78         \\
\textsc{DeepWalk}   & 0.28 (5.1e-3)  & 0.69 (7.6e-3)  & 165.9 (55.6) & 104.91 & 1049.89        \\ \hline
\end{tabular}
\label{table:sphere_experiment_individual}
\end{table}

\begin{table}[ht!]
\centering
\caption{Numerical results running models individually on the ERA5 Temperature dataset $\lambda_t$ and $T_t$ correspond to the average time and total time to run the models in seconds. The standard deviations are displayed within parenthesis. }
\begin{tabular}{l|ccccc}
\hline
       & $Q_{local}$ & $Q_{global}$ & $K_{\max}$ &  $\lambda_t$ & $T_t$  \\ \hline
\textsc{PCA}  & {0.89} (1.6e-2)  & \textbf{0.99} (1.1e-3)  & 11.1 (2.7) & \textbf{0.09}  & \textbf{1.06}          \\
\textsc{MDS}    & \textbf{0.89} (1.1e-2)  & 0.99 (6.1e-4)  & 11.6 (1.2) & 0.42 & 4.4        \\
\textsc{LLE}    & 0.53 (3.3e-2)  & 0.78 (1.8e-2) & 12.6 (1.8) & 0.25 & 2.60         \\
\textsc{Isomap} & 0.83 (1.0e-2)  & 0.99 (1.2e-2) & 11.4 (1.2) & 0.18 & 1.39         \\
\textsc{SE}  & 0.62 (2.4e-2)  & 0.83 (1.1e-2) & 23.3 (3.3) & 0.12 & 1.39         \\
\textsc{GCN}  & 0.29 (5.4e-2)  & 0.70 (3.4e-2)  & 42.3 (8.3) & 13.3 & 133.12          \\
\textsc{GS}   & 0.61 (4.6e-2)  & 0.91 (8.8e-3)  & 31.1 (6.0) & 35.38 & 353.98        \\
\textsc{GAT}    &  0.53 (3.0e-2)  & 0.87 (1.3e-2)  & 31.8 (1.99) & 65.00  & 650.13         \\
\textsc{DeepWalk}   & 0.64 (5.2e-2)  & 0.90 (2.9e-3)  & 29.1 (4.89) & 29.4 & 293.61        \\ \hline
\end{tabular}
\label{table:ERA5_experiment_individual}
\end{table}

To compare the computational time and performance of the automated pipeline compared to running the models individually we used $n_{iter}=10$ using random number seeds for each of the dynamical datasets and documented the embedding quality of the best performing model in addition to the run time. For the fKdV data, \textsc{PCA} yielded the best embedding quality yielding $Q_{local}=0.86$ and $Q_{global}=0.99$ where the pipeline took 59.95 seconds. For the KS data, \textsc{PCA} yielded the best embedding quality yielding $Q_{local}=0.99$ and $Q_{global}=0.99$ where the pipeline took 87.21 seconds. Finally for the SG data, \textsc{PCA} yielded the best embedding quality yielding $Q_{local}=0.96$ and $Q_{global}=0.99$ where the pipeline took 64.99 seconds. For the static Swiss Roll dataset, our automated pipeline yielded \textsc{Isomap} as the best performing model where $Q_{local}=0.87$ and $Q_{global}=0.91$. The pipeline took 36.01 seconds to run.
All of the experimental results presented in the preceding tables were run on CPUs only.

\newpage
\section{Additional Experimental Results}\label{apx:exp}
Below are the three dimensional representations corresponding to the results presented in Table\ref{tab:results_table} for each equation and model.
\begin{figure}[ph]
    \centering
\includegraphics[width=0.7\textwidth]{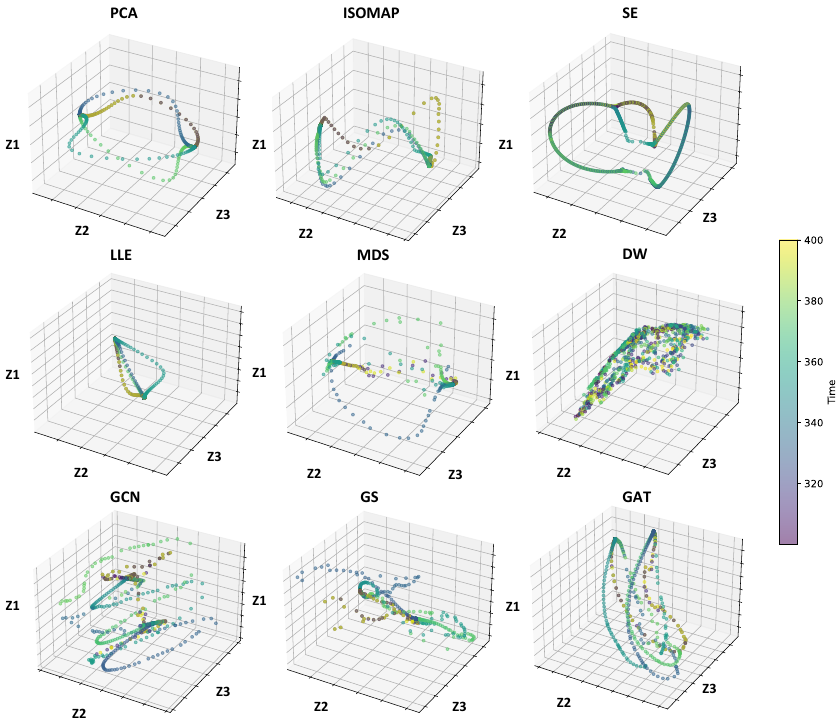}
    \caption{Three-dimensional visualizations of the reduced order dynamics for the KS equation for all of the techniques presented in this study. We see that classical methods outperform deep learning based methods. Among the classical methods, the best-performing ones are global embedding methods. }
    \label{fig:app_ks}
\end{figure}

\begin{figure}[ph]
    \centering
\includegraphics[width=0.7\textwidth]{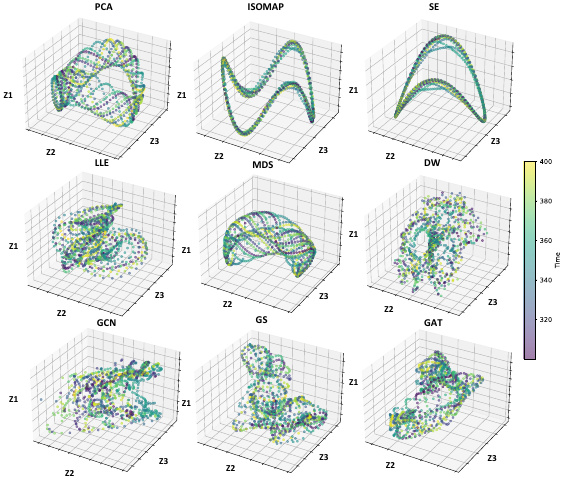}
    \caption{Three-dimensional visualizations of the reduced order dynamics for the fKdV equation for all of the techniques presented in this study. We see that classical methods outperform deep learning based methods. Among the classical methods, the best-performing ones are global embedding methods. }
    \label{fig:app_fkdv}
\end{figure}

\begin{figure}[ph]
    \centering
\includegraphics[width=0.7\textwidth]{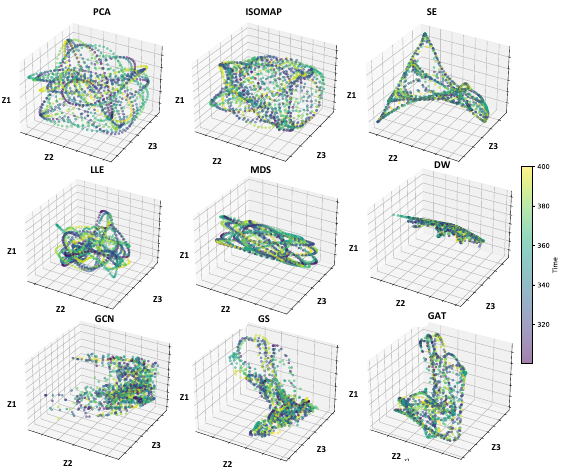}
    \caption{Three-dimensional visualizations of the reduced order dynamics for the SG equation for all of the techniques presented in this study. We see that classical methods outperform deep learning based methods. Among the classical methods, the best-performing ones are global embedding methods. }
    \label{fig:app_sg}
\end{figure}



\end{document}